\documentclass[sigconf]{acmart}

\usepackage{hyperref}
\usepackage{url}

\usepackage{graphicx}
\usepackage{subcaption}
\usepackage{multirow}
\usepackage{multicol}
\usepackage{booktabs}

\newcommand{\R}{\mathbb{R}}   
\newcommand{\vect}[1]{\mathbf{#1}}   
\newcommand{\mat}[1]{\mathbf{#1}}    
\newcommand{\appropto}{\mathrel{\vcenter{
  \offinterlineskip\halign{\hfil$##$\cr
    \propto\cr\noalign{\kern2pt}\sim\cr\noalign{\kern-2pt}}}}}
    
\definecolor{demphcolor}{RGB}{144,144,144}
\newcommand{\demph}[1]{\textcolor{demphcolor}{#1}}

\AtBeginDocument{%
  \providecommand\BibTeX{{%
    \normalfont B\kern-0.5em{\scshape i\kern-0.25em b}\kern-0.8em\TeX}}}

\copyrightyear{2022}
\acmYear{2022}
\setcopyright{acmcopyright}\acmConference[MM '22]{Proceedings of the 30th ACM International Conference on Multimedia}{October 10--14, 2022}{Lisboa, Portugal}
\acmBooktitle{Proceedings of the 30th ACM International Conference on Multimedia (MM '22), October 10--14, 2022, Lisboa, Portugal}
\acmPrice{15.00}
\acmDOI{10.1145/3503161.3548112}
\acmISBN{978-1-4503-9203-7/22/10}

\begin{document}
\fancyhead{}

\title[LayoutLMv3: Pre-training for Document AI with Unified Text and Image Masking]{LayoutLMv3: Pre-training for Document AI \\ with Unified Text and Image Masking}


\author{Yupan Huang}
\authornote{Contribution during internship at Microsoft Research. Corresponding authors: Lei Cui and Furu Wei.}
\affiliation{\institution{Sun Yat-sen University}
\country{}}
\email{huangyp28@mail2.sysu.edu.cn}

\author{Tengchao Lv}
\affiliation{\institution{Microsoft Research Asia}
\country{}}
\email{tengchaolv@microsoft.com}

\author{Lei Cui}
\affiliation{\institution{Microsoft Research Asia}
\country{}}
\email{lecu@microsoft.com}

\author{Yutong Lu}
\affiliation{\institution{Sun Yat-sen University}
\country{}}
\email{luyutong@mail.sysu.edu.cn}

\author{Furu Wei}
\affiliation{\institution{Microsoft Research Asia}
\country{}}
\email{fuwei@microsoft.com}
\renewcommand{\shortauthors}{Huang, et al.}

\begin{abstract}
Self-supervised pre-training techniques have achieved remarkable progress in Document AI. Most multimodal pre-trained models use a masked language modeling objective to learn bidirectional representations on the text modality, but they differ in pre-training objectives for the image modality. This discrepancy adds difficulty to multimodal representation learning. In this paper, we propose \textbf{LayoutLMv3} to pre-train multimodal Transformers for Document AI with unified text and image masking. Additionally, LayoutLMv3 is pre-trained with a word-patch alignment objective to learn cross-modal alignment by predicting whether the corresponding image patch of a text word is masked. The simple unified architecture and training objectives make LayoutLMv3 a general-purpose pre-trained model for both text-centric and image-centric Document AI tasks. Experimental results show that LayoutLMv3 achieves state-of-the-art performance not only in text-centric tasks, including form understanding, receipt understanding, and document visual question answering, but also in image-centric tasks such as document image classification and document layout analysis. The code and models are publicly available at \url{https://aka.ms/layoutlmv3}.
\end{abstract}



\maketitle


\begin{figure}[t]
\centering
    \begin{subfigure}[b]{0.45\linewidth}
        \fbox{\includegraphics[width=\linewidth]{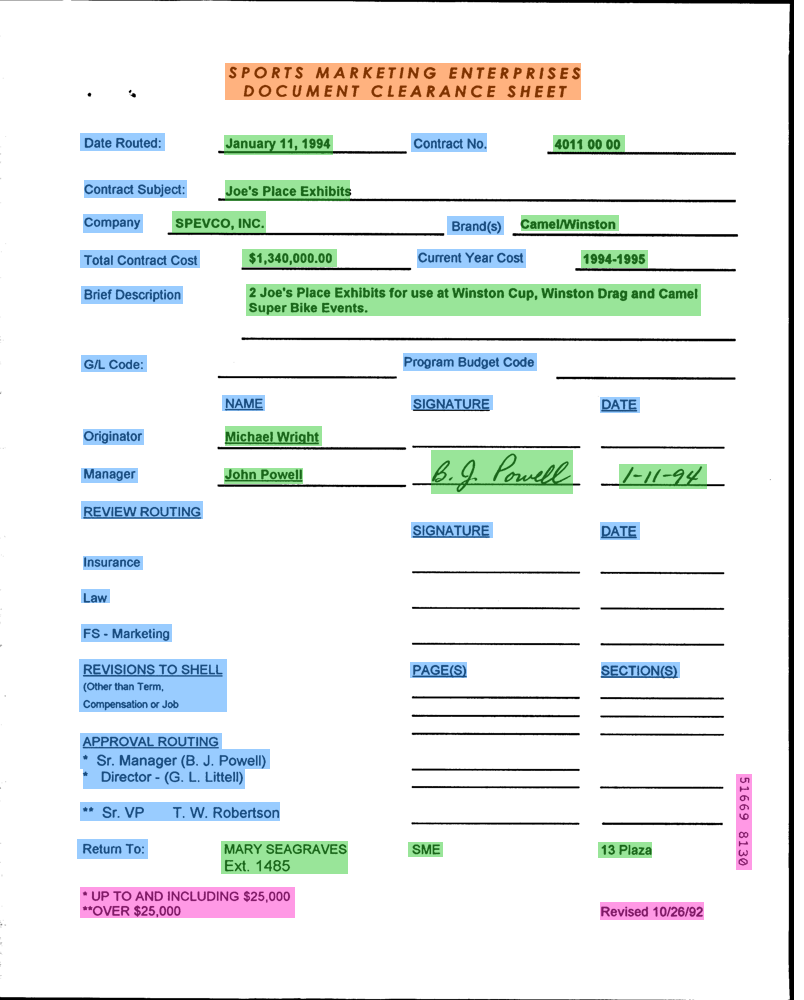}}
        \caption{Text-centric form understanding on FUNSD}
        \label{fig:1a}
    \end{subfigure}
    \qquad
    \begin{subfigure}[b]{0.44\linewidth}
        \fbox{\includegraphics[width=0.9\linewidth]{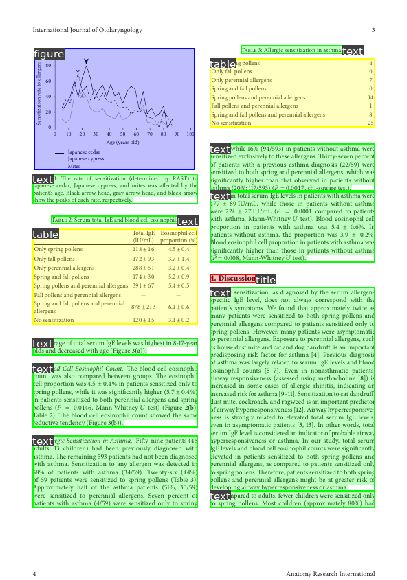}}
        \caption{Image-centric layout analysis on PubLayNet}
        \label{fig:1b}
    \end{subfigure}
    \caption{Examples of Document AI Tasks.}
    \label{fig:1}
\end{figure} 

\begin{figure}[t]
    \centering
    \includegraphics[width=1\linewidth]{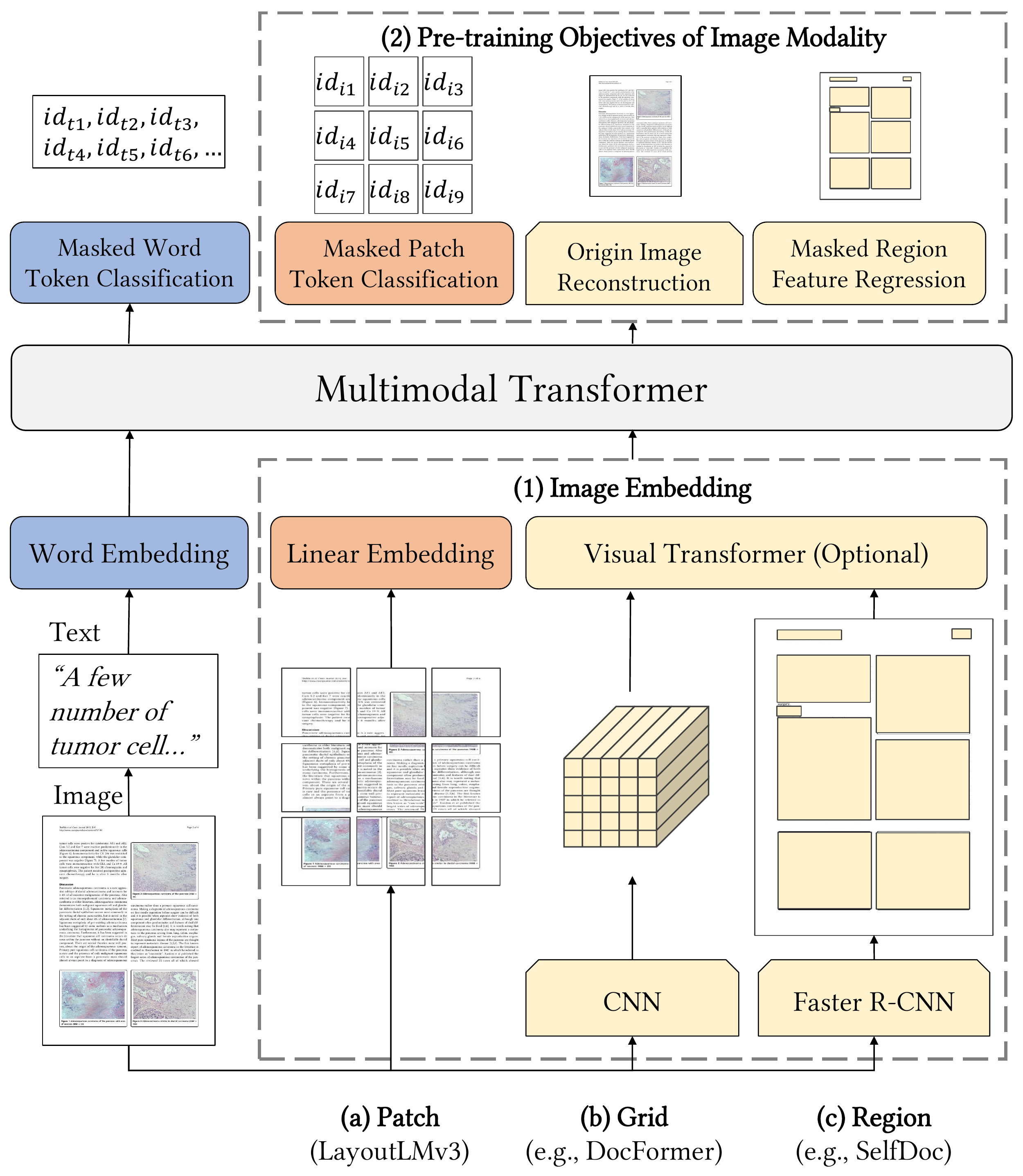}
    \caption{\textbf{Comparisons with existing works} (e.g., DocFormer \protect\cite{Appalaraju_2021_ICCV} and SelfDoc \protect\cite{li2021selfdoc}) on
    \textbf{(1) image embedding}: our LayoutLMv3 uses linear patches to reduce the computational bottleneck of CNNs and eliminate the need for region supervision in training object detectors;
    \textbf{(2) pre-training objectives on image modality}: our LayoutLMv3 learns to reconstruct discrete image tokens of masked patches instead of raw pixels or region features to capture high-level layout structures rather than noisy details.
    } \label{fig:intro}
\end{figure}

\section{Introduction}

In recent years, pre-training techniques have been making waves in the Document AI community by achieving remarkable progress on document understanding tasks~\cite{xu2020layoutlm,xu-etal-2021-layoutlmv2,xu2021layoutxlm,pramanik2020towards,garncarek2021lambert,hong2022bros,Powalski2021GoingFB,wu2021lampret,Li2021StructuralLMSP,li2021selfdoc,Appalaraju_2021_ICCV,li2021structext,gu2021unidoc,wang2022LiLT,gu2022xylayoutlm,lee2022formnet}. As shown in Figure~\ref{fig:1}, a pre-trained Document AI model can parse layout and extract key information for various documents such as scanned forms and academic papers, which is important for industrial applications and academic research~\cite{cui2021document}.

Self-supervised pre-training techniques have made rapid progress in representation learning due to their successful applications of reconstructive pre-training objectives.
In NLP research, BERT firstly proposed ``masked language modeling'' (MLM) to learn bidirectional representations by predicting the original vocabulary id of a randomly masked word token based on its context~\cite{devlin2019bert}.
Whereas most performant multimodal pre-trained Document AI models use the MLM proposed by BERT for text modality, they differ in pre-training objectives for image modality as depicted in Figure~\ref{fig:intro}.
For example, DocFormer learns to reconstruct image pixels through a CNN decoder~\cite{Appalaraju_2021_ICCV}, which tends to learn noisy details rather than high-level structures such as document layouts~\cite{salimans2017pixelcnn++,ramesh2021zero}.
SelfDoc proposes to regress masked region features~\cite{li2021selfdoc}, which is noisier and harder to learn than classifying discrete features in a smaller vocabulary~\cite{cho2020x,huang2021unifying}.
The different granularities of image (e.g., dense image pixels or contiguous region features) and text (i.e., discrete tokens) objectives further add difficulty to cross-modal alignment learning, which is essential to multimodal representation learning.

\begin{figure*}[t]
    \centering
    \includegraphics[width=0.9\linewidth]{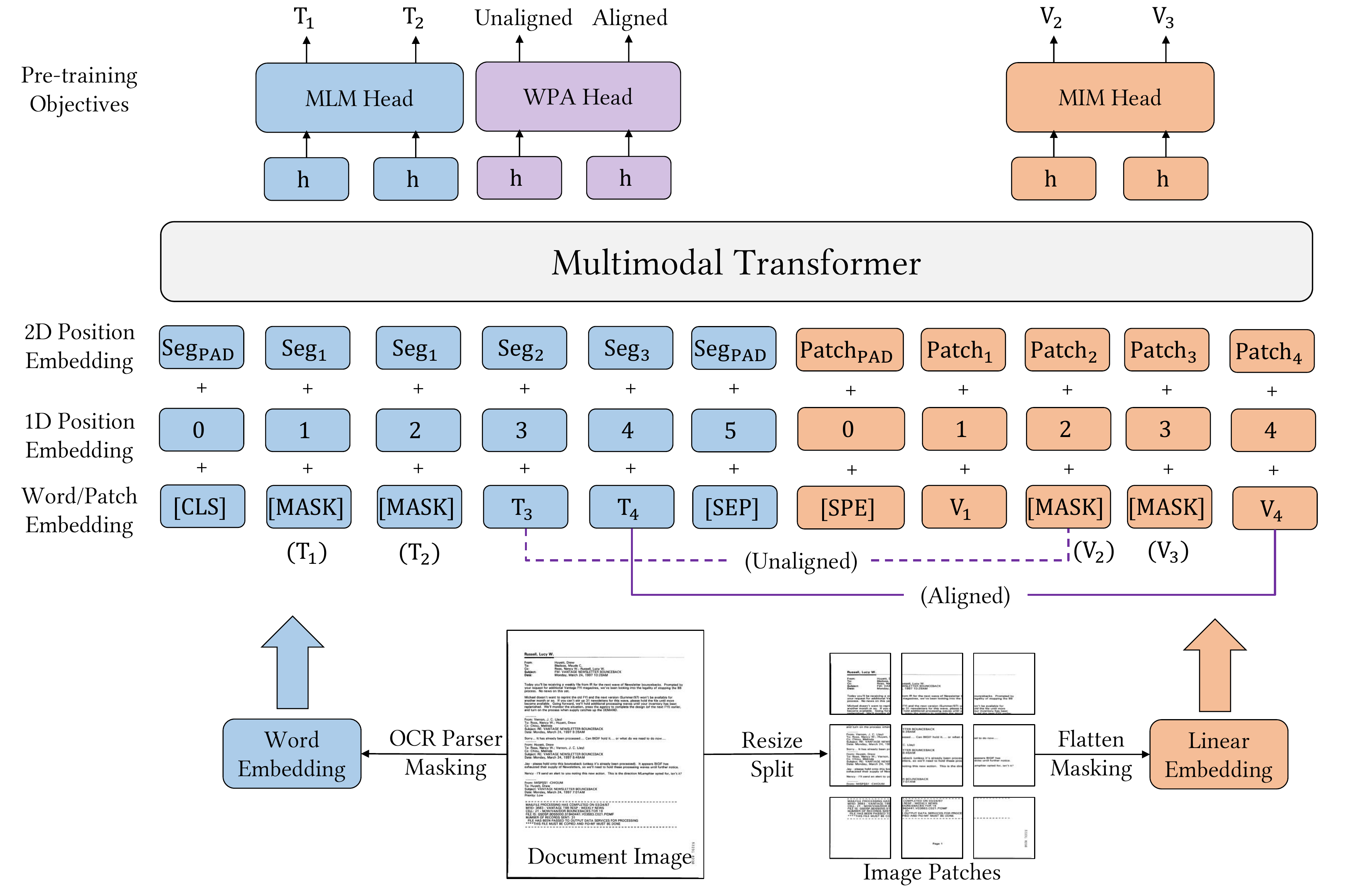}
    \caption{
    \textbf{The architecture and pre-training objectives of LayoutLMv3.}
    LayoutLMv3 is a pre-trained multimodal Transformer for Document AI with unified text and image masking objectives.
    Given an input document image and its corresponding text and layout position information, the model takes the linear projection of patches and word tokens as inputs and encodes them into contextualized vector representations.
    LayoutLMv3 is pre-trained with discrete token reconstructive objectives of Masked Language Modeling (MLM) and Masked Image Modeling (MIM).
    Additionally, LayoutLMv3 is pre-trained with a Word-Patch Alignment (WPA) objective to learn cross-modal alignment by predicting whether the corresponding image patch of a text word is masked. ``Seg'' denotes segment-level positions. ``[CLS]'', ``[MASK]'', ``[SEP]'' and ``[SPE]'' are special tokens.
    } \label{fig:architecture}
\end{figure*}

To overcome the discrepancy in pre-training objectives of text and image modalities and facilitate multimodal representation learning, we propose \textbf{LayoutLMv3} to pre-train multimodal Transformers for Document AI with unified text and image masking objectives MLM and MIM.
As shown in Figure~\ref{fig:architecture}, LayoutLMv3 learns to reconstruct masked word tokens of the text modality and symmetrically reconstruct masked patch tokens of the image modality.
Inspired by DALL-E~\cite{ramesh2021zero} and BEiT~\cite{bao2022beit}, we obtain the target image tokens from latent codes of a discrete VAE.
For documents, each text word corresponds to an image patch. To learn this cross-modal alignment, we propose a Word-Patch Alignment (WPA) objective to predict whether the corresponding image patch of a text word is masked.

Inspired by ViT~\cite{dosovitskiy2020vit} and ViLT~\cite{kim2021vilt}, LayoutLMv3 directly leverages raw image patches from document images without complex pre-processing steps such as page object detection.
LayoutLMv3 jointly learns image, text and multimodal representations in a Transformer model with unified MLM, MIM and WPA objectives.
This makes LayoutLMv3 the first multimodal pre-trained Document AI model without CNNs for image embeddings, which significantly saves parameters and gets rid of region annotations.
The simple unified architecture and objectives make LayoutLMv3 a general-purpose pre-trained model for both text-centric tasks and image-centric Document AI tasks.

We evaluated pre-trained LayoutLMv3 models across five public benchmarks, including text-centric benchmarks: FUNSD~\cite{jaume2019funsd} for form understanding, CORD~\cite{park2019cord} for receipt understanding, DocVQA~\cite{mathew2021docvqa} for document visual question answering, and image-centric benchmarks: RVL-CDIP~\cite{harley2015icdar} for document image classification, PubLayNet~\cite{zhong2019publaynet} for document layout analysis. 
Experiment results demonstrate that LayoutLMv3 achieves state-of-the-art performance on these benchmarks with parameter efficiency. 
Furthermore, LayoutLMv3 is easy to reproduce for its simple and neat architecture and pre-training objectives.

Our contributions are summarized as follows:
\begin{itemize}
	\item LayoutLMv3 is the first multimodal model in Document AI that does not rely on a pre-trained CNN or Faster R-CNN backbone to extract visual features, which significantly saves parameters and eliminates region annotations.
	\item LayoutLMv3 mitigates the discrepancy between text and image multimodal representation learning with unified discrete token reconstructive objectives MLM and MIM. We further propose a Word-Patch Alignment (WPA) objective to facilitate cross-modal alignment learning.
	\item LayoutLMv3 is a general-purpose model for both text-centric and image-centric Document AI tasks. For the first time, we demonstrate the generality of multimodal Transformers to vision tasks in Document AI.
	\item Experimental results show that LayoutLMv3 achieves state-of-the-art performance in text-centric tasks and image-centric tasks in Document AI. The code and models are publicly available at \url{https://aka.ms/layoutlmv3}.
\end{itemize}

\section{LayoutLMv3}
Figure~\ref{fig:architecture} gives an overview of the LayoutLMv3.

\subsection{Model Architecture}
LayoutLMv3 applies a unified text-image multimodal Transformer to learn cross-modal representations.
The Transformer has a multi-layer architecture and each layer mainly consists of multi-head self-attention and 
position-wise fully connected 
feed-forward networks~\cite{vaswani2017attention}.
The input of Transformer is a concatenation of text embedding $\mat{Y}=\vect{y}_{1:L}$ and image embedding $\mat{X}=\vect{x}_{1:M}$ sequences, where $L$ and $M$ are sequence lengths for text and image respectively.
Through the Transformer, the last layer outputs text-and-image contextual representations.

\noindent \textbf{Text Embedding.}
Text embedding is a combination of word embeddings and position embeddings.
We pre-processed document images with an off-the-shelf OCR toolkit to obtain textual content and corresponding 2D position information.
We initialize \textit{the word embeddings} with a word embedding matrix from a pre-trained model RoBERTa~\cite{liu2019roberta}.
\textit{The position embeddings} include 1D position and 2D layout position embeddings, where the 1D position refers to the index of tokens within the text sequence, and the \textbf{2D layout position} refers to the bounding box coordinates of the text sequence.
Following the LayoutLM, we normalize all coordinates by the size of images, and use embedding layers to embed x-axis, y-axis, width and height features separately~\cite{xu2020layoutlm}.
The LayoutLM and LayoutLMv2 adopt word-level layout positions, where each word has its positions.
Instead, we adopt segment-level layout positions that words in a segment share the same 2D position since the words usually express the same semantic meaning~\cite{Li2021StructuralLMSP}.

\noindent \textbf{Image Embedding.}
Existing multimodal models in Document AI either extract CNN grid features~\cite{xu-etal-2021-layoutlmv2,Appalaraju_2021_ICCV} or rely on an object detector like Faster R-CNN~\cite{Ren2015FasterRT} to extract region features~\cite{xu2020layoutlm,Powalski2021GoingFB,li2021selfdoc,gu2021unidoc} for image embeddings, which accounts for heavy computation bottleneck or require region supervision.
Inspired by ViT~\cite{dosovitskiy2020vit} and ViLT~\cite{kim2021vilt}, we represent document images with linear projection features of image patches before feeding them into the multimodal Transformer.
Specifically, we resize a document image into $H\times W$ and denote the image with $\vect{I} \in \R^{C \times H \times W}$, where $C$, $H$ and $W$ are the channel size, width and height of the image respectively.
We then split the image into a sequence of uniform $P\times P$ patches, linearly project the image patches to $D$ dimensions and flatten them into a sequence of vectors, which length is $M={HW}/{P^2}$.
Then we add learnable 1D position embeddings to each patch since we have not observed improvements from using 2D position embeddings in our preliminary experiments.
LayoutLMv3 is the first multimodal model in Document AI that does not rely on CNNs to extract image features, which is vital to Document AI models to reduce parameters or remove complex pre-processing steps.

We insert semantic 1D relative position and spatial 2D relative position as bias terms in self-attention networks for text and image modalities following LayoutLMv2\cite{xu-etal-2021-layoutlmv2}.

\subsection{Pre-training Objectives}
LayoutLMv3 is pre-trained with the MLM, MIM, and WPA objectives to learn multimodal representation in a self-supervised learning manner.
Full pre-training objectives of LayoutLMv3 is defined as $L = L_{MLM} + L_{MIM} + L_{WPA}$.

\noindent \textbf{Objective I: Masked Language Modeling (MLM).}
For the language side, our MLM is inspired by the masked language modeling in BERT~\cite{devlin2019bert} and masked visual-language modeling in LayoutLM~\cite{xu2020layoutlm} and LayoutLMv2~\cite{xu-etal-2021-layoutlmv2}.
We mask 30\% of text tokens with a span masking strategy with span lengths drawn from a Poisson distribution ($\lambda=3$)~\cite{lewis2020bart,joshi2020spanbert}.
The pre-training objective is to maximize the log-likelihood of the correct masked text tokens $\vect{y}_l$ based on the contextual representations of corrupted sequences of image tokens $\mat{X}^{M'}$ and text tokens $\mat{Y}^{L'}$, where $M'$ and $L'$ represent the masked positions.
We denote parameters of the Transformer model with $\theta$ and minimize the subsequent cross-entropy loss:
\begin{align}
L_{MLM}\left(\theta\right) &= -\sum_{l=1}^{L'} \log p_{\theta}\left(\vect{y}_\ell \mid \mat{X}^{M'}, \mat{Y}^{L'}\right)
\end{align}
As we keep the layout information unchanged, this objective facilitates the model to learn the correspondence between layout information and text and image context.

\noindent \textbf{Objective II: Masked Image Modeling (MIM).}
To encourage the model to interpret visual content from contextual text and image representations, we adapt the MIM pre-training objective in BEiT~\cite{bao2022beit} to our multimodal Transformer model.
The MIM objective is a symmetry to the MLM objective, that we randomly mask a percentage of about 40\% image tokens with the blockwise masking strategy~\cite{bao2022beit}.
The MIM objective is driven by a cross-entropy loss to reconstruct the masked image tokens $\vect{x}_m$ under the context of their surrounding text and image tokens.
\begin{align}
L_{MIM}\left(\theta\right) &= -\sum_{m=1}^{M'}\log p_{\theta}\left(\vect{x}_m \mid \mat{X}^{M'}, \mat{Y}^{L'}\right)
\end{align}
The labels of image tokens come from an image tokenizer, which can transform dense image pixels into discrete tokens according to a visual vocabulary~\cite{ramesh2021zero}.
Thus MIM facilitates learning high-level layout structures rather than noisy low-level details.

\noindent \textbf{Objective III: Word-Patch Alignment (WPA).}
For documents, each text word corresponds to an image patch. As we randomly mask text and image tokens with MLM and MIM respectively, there is no explicit alignment learning between text and image modalities.
We thus propose a WPA objective to learn a fine-grained alignment between text words and image patches.
The WPA objective is to predict whether the corresponding image patches of a text word are masked.
Specifically, we assign an \emph{aligned} label to an \emph{unmasked} text token when its corresponding image tokens are also unmasked.
Otherwise, we assign an \emph{unaligned} label.
We exclude the \emph{masked} text tokens when calculating WPA loss to prevent the model from learning a correspondence between masked text words and image patches.
We use a two-layer MLP head that inputs contextual text and image and outputs the binary {aligned}/{unaligned} labels with a binary cross-entropy loss:
\begin{align}
L_{WPA}\left(\theta\right) &= -\sum_{\ell=1}^{L-L'} \log p_{\theta}\left(\vect{z}_\ell \mid \mat{X}^{M'}, \mat{Y}^{L'}\right),
\end{align}
where $L-L'$ is the number of unmasked text tokens, $\vect{z}_\ell$ is the binary label of language token in the $\ell$ position.

\begin{table*}[t]
    \centering
    \caption{\textbf{Comparison with existing published models} on the CORD \protect\cite{park2019cord}, FUNSD \protect\cite{jaume2019funsd}, RVL-CDIP \protect\cite{harley2015icdar}, and DocVQA \protect\cite{mathew2021docvqa} datasets.
    ``T/L/I'' denotes ``text/layout/image'' modality. ``R/G/P'' denotes ``region/grid/patch'' image embedding.
    We multiply all values by a hundred for better readability.
    $^\dagger$In the UDoc paper \protect\cite{gu2021unidoc}, the CORD splits are 626/247 receipts for training/test instead of the official 800/100 training/test receipts adopted by other works. Thus the score$^\dagger$ is not directly comparable to other scores.
    Models denoted with $^\ddagger$ use more data to train DocVQA and are expected to score higher. For example, TILT introduces one more supervised training stage on more QA datasets \protect\cite{Powalski2021GoingFB}. StructuralLM additionally uses the validation set in training \protect\cite{Li2021StructuralLMSP}.
    }
    \label{tab:sota}
    \begin{tabular}{llllcccc}
    \toprule
    \multirow{2}{*}{\bf Model} & \multirow{2}{*}{\bf Parameters} & \multirow{2}{*}{\bf Modality} & \multirow{2}{*}{\bf Image Embedding} & \bf FUNSD & \bf CORD & \bf RVL-CDIP  & \bf DocVQA  \\
     & & & & \bf F1$\uparrow$  & \bf F1$\uparrow$ & \bf Accuracy$\uparrow$ & \bf ANLS$\uparrow$ \\
     \midrule
     $\textrm{BERT}_{\rm BASE}$~\cite{devlin2019bert} & 110M & T & None & 60.26 & 89.68 &  89.81 & 63.72\\
     $\textrm{RoBERTa}_{\rm BASE}$~\cite{liu2019roberta} & 125M & T & None & 66.48 & 93.54 &  90.06 & 66.42\\
     $\textrm{BROS}_{\rm BASE}$~\cite{hong2022bros} & 110M & T+L & None & 83.05 & 95.73 &  - & -\\
     $\textrm{LiLT}_{\rm BASE}$~\cite{wang2022LiLT} & - & T+L & None & 88.41 & 96.07 & 95.68* & - \\
     $\textrm{LayoutLM}_{\rm BASE}$~\cite{xu2020layoutlm} & 160M & T+L+I (R) & ResNet-101 (fine-tune) & 79.27 & - & 94.42 & -\\
     $\textrm{SelfDoc}$~\cite{li2021selfdoc} & - & T+L+I (R) & ResNeXt-101 & 83.36 & - & 92.81 & -\\
     $\textrm{UDoc}$~\cite{gu2021unidoc} & 272M & T+L+I (R) & ResNet-50 & 87.93 & \demph{98.94}$^\dagger$  & 95.05 & -\\
     $\textrm{TILT}_{\rm BASE}$~\cite{Powalski2021GoingFB} & 230M & T+L+I (R) & U-Net & - & 95.11 & 95.25 & \demph{83.92}$^\ddagger$\\
     $\textrm{XYLayoutLM}_{\rm BASE}$~\cite{gu2022xylayoutlm} & - & T+L+I (G) & ResNeXt-101 & 83.35 & - & - & -\\
     $\textrm{LayoutLMv2}_{\rm BASE}$~\cite{xu-etal-2021-layoutlmv2} & 200M & T+L+I (G) & ResNeXt101-FPN & 82.76 & 94.95 & 95.25 & 78.08\\
     $\textrm{DocFormer}_{\rm BASE}$~\cite{Appalaraju_2021_ICCV} & 183M & T+L+I (G) & ResNet-50 & 83.34 & 96.33 & \textbf{96.17} & - \\
     \bf $\textrm{LayoutLMv3}_{\rm BASE}$ (Ours) & 133M & T+L+I (P) & Linear & \textbf{90.29} & \textbf{96.56} & 95.44 & \textbf{78.76} \\
     \midrule
     $\textrm{BERT}_{\rm LARGE}$~\cite{devlin2019bert} & 340M & T& None & 65.63 & 90.25 & 89.92 & 67.45\\
     $\textrm{RoBERTa}_{\rm LARGE}$~\cite{liu2019roberta} & 355M & T& None & 70.72 & 93.80 &  90.11 & 69.52 \\
     $\textrm{LayoutLM}_{\rm LARGE}$~\cite{xu2020layoutlm} & 343M & T+L& None & 77.89 & - & 91.90 & - \\
     $\textrm{BROS}_{\rm LARGE}$~\cite{hong2022bros} & 340M & T+L& None & 84.52 & 97.40 &  - & -\\
     $\textrm{StructuralLM}_{\rm LARGE}$~\cite{Li2021StructuralLMSP} & 355M & T+L& None & 85.14 & - & \textbf{96.08} & \demph{83.94}$^\ddagger$ \\
     $\textrm{FormNet}$~\cite{lee2022formnet} & 217M & T+L & None & 84.69 & - & - & - \\
     $\textrm{FormNet}$~\cite{lee2022formnet} & 345M & T+L & None & - & 97.28 & - & - \\
     $\textrm{TILT}_{\rm LARGE}$~\cite{Powalski2021GoingFB} & 780M & T+L+I (R) & U-Net & - & 96.33 & 95.52 & \demph{87.05}$^\ddagger$\\
     $\textrm{LayoutLMv2}_{\rm LARGE}$~\cite{xu-etal-2021-layoutlmv2} & 426M & T+L+I (G) & ResNeXt101-FPN & 84.20 & 96.01 & 95.64 & \textbf{83.48} \\
     $\textrm{DocFormer}_{\rm LARGE}$~\cite{Appalaraju_2021_ICCV} & 536M & T+L+I (G) & ResNet-50 & 84.55 & 96.99 & 95.50 & - \\
     \bf $\textrm{LayoutLMv3}_{\rm LARGE}$ (Ours) & 368M & T+L+I (P) & Linear & \textbf{92.08} & \textbf{97.46} & {95.93} & 83.37 \\
     \bottomrule
    \multicolumn{7}{l}{\footnotesize 
    * LiLT uses image features with ResNeXt101-FPN backbone in fine-tuning RVL-CDIP.
    }
    \end{tabular}
\end{table*}

\section{Experiments}

\subsection{Model Configurations}
The network architecture of LayoutLMv3 follows that of LayoutLM~\cite{xu2020layoutlm} and LayoutLMv2~\cite{xu-etal-2021-layoutlmv2} for a fair comparison.
We use base and large model sizes for LayoutLMv3.
$\mathrm{LayoutLMv3_{BASE}}$ adopts a 12-layer Transformer encoder with 12-head self-attention, hidden size of $D=768$, and 3,072 intermediate size of feed-forward networks.
$\mathrm{LayoutLMv3_{LARGE}}$ adopts a 24-layer Transformer encoder with 16-head self-attention, hidden size of $D=1,024$, and 4,096 intermediate size of feed-forward networks.
To pre-process the text input, we tokenize the text sequence with Byte-Pair Encoding (BPE)~\cite{sennrich2016neural} with a maximum sequence length $L=512$.
We add a [CLS] and a [SEP] token at the beginning and end of each text sequence. When the length of the text sequence is shorter than $L$, we append [PAD] tokens to it. The bounding box coordinates of these special tokens are all zeros.
The parameters for image embedding are $C\times H\times W = 3\times 224 \times 224$, $P=16$, $M=196$.

We adopt distributed and mixed-precision training to reduce memory costs and speed up training procedures.
We also use a gradient accumulation mechanism to split the batch of samples into several mini-batches to overcome memory constraints for large batch sizes.
We further use a gradient checkpointing technique for document layout analysis to reduce memory costs.
To stabilize training, we follow CogView~\cite{ding2021cogview} to change the computation of attention to $\textrm{softmax}\left(\frac{\mat{Q}^T\mat{K}}{\sqrt{d}}\right)  =\textrm{softmax}\left(\left(\frac{\mat{Q}^T}{\alpha\sqrt{d}}\mat{K} - \max\left(\frac{\mat{Q}^T}{\alpha\sqrt{d}}\mat{K}\right)\right)\times \alpha\right)$, where $\alpha$ is 32.

\subsection{Pre-training LayoutLMv3}
To learn a universal representation for various document tasks, we pre-train LayoutLMv3 on a large IIT-CDIP dataset. 
The \textbf{IIT-CDIP Test Collection 1.0} is a large-scale scanned document image dataset, which contains about 11 million document images and can split into 42 million pages~\cite{Lewis:2006:BTC:1148170.1148307}.
We only use 11 million of them to train LayoutLMv3.
We do not do image augmentation following LayoutLM models~\cite{xu2020layoutlm, xu-etal-2021-layoutlmv2}. 
For the multimodal Transformer encoder along with the text embedding layer, LayoutLMv3 is initialized from the pre-trained weights of RoBERTa~\cite{liu2019roberta}.
Our image tokenizer is initialized from a pre-trained image tokenizer in DiT, a self-supervised pre-trained document image Transformer model~\cite{li2022dit}.
The vocabulary size of image tokens is 8,192.
We randomly initialized the rest model parameters.
We pre-train LayoutLMv3 using Adam optimizer~\cite{kingma2014adam} with a batch size of 2,048 for 500,000 steps.
We use a weight decay of $1e-2$, and ($\beta$1, $\beta$2) = (0.9, 0.98).
For the $\mathrm{LayoutLMv3_{BASE}}$ model, we use a learning rate of $1e-4$, and we linearly warm up the learning rate over the first 4.8\% steps.
For $\mathrm{LayoutLMv3_{LARGE}}$, the learning rate and warm-up ratio are $5e-5$ and 10\%, respectively.

\subsection{Fine-tuning on Multimodal Tasks}
We compare LayoutLMv3 with typical self-supervised pre-training approaches and categorize them by their pre-training modalities.
\begin{itemize}
    \item \textbf{[T] text modality}: BERT~\cite{devlin2019bert} and RoBERTa~\cite{liu2019roberta} are typical pre-trained language models which only use text information with Transformer architecture. 
    We use FUNSD and RVL-CDIP results of the RoBERTa from LayoutLM~\cite{xu2020layoutlm} and results of BERT from LayoutLMv2~\cite{xu-etal-2021-layoutlmv2}. We reproduce and report the CORD and DocVQA results of the RoBERTa.
    \item \textbf{[T+L] text and layout modalities}: LayoutLM incorporates layout information by adding word-level spatial embeddings to embeddings of BERT~\cite{xu2020layoutlm}. StructuralLM leverages segment-level layout information~\cite{Li2021StructuralLMSP}. BROS encodes relative layout positions~\cite{hong2022bros}. LILT fine-tunes on different languages with pre-trained textual models~\cite{wang2022LiLT}.
    FormNet leverages the spatial relationship between tokens in a form~\cite{lee2022formnet}.
    \item \textbf{[T+L+I (R)] text, layout and image modalities with Faster R-CNN region features}: 
	This line of works extract image region features from RoI heads in the Faster R-CNN model~\cite{Ren2015FasterRT}. Among them, LayoutLM~\cite{xu2020layoutlm} and TILT~\cite{Powalski2021GoingFB} use OCR words' bounding box to serve as region proposals and add the region features to corresponding text embeddings.
	SelfDoc~\cite{li2021selfdoc} and UDoc~\cite{gu2021unidoc} use document object proposals and concatenate region features with text embeddings.
    \item \textbf{[T+L+I (G)] text, layout and image modalities with CNN grid features}: LayoutLMv2~\cite{xu-etal-2021-layoutlmv2} and DocFormer~\cite{Appalaraju_2021_ICCV} extract image grid features with a CNN backbone without object detection. XYLayoutLM~\cite{gu2022xylayoutlm} adopts the architecture of LayoutLMv2 and improves layout representation.
    \item \textbf{[T+L+I (P)] text, layout, and image modalities with linear patch features}: LayoutLMv3 replaces CNN backbones with simple linear embedding to encode image patches.
\end{itemize}

\begin{table*}[t]
    \centering
    \caption{\textbf{Document layout analysis} mAP @ IOU [0.50:0.95] on PubLayNet validation set. All models use only information from the vision modality.
    LayoutLMv3 outperforms the compared ResNets \protect\cite{zhong2019publaynet, gu2021unidoc} and vision Transformer \protect\cite{li2022dit} backbones.
    }
    \label{tab:publaynet}
    \begin{tabular}{ccccccccc}
        \toprule
        \multicolumn{1}{c}{\bf Model} & \bf Framework & \bf Backbone & \bf Text & \bf Title & \bf List & \bf Table & \bf Figure & \bf Overall \\
        \midrule
        PubLayNet\cite{zhong2019publaynet} & Mask R-CNN & ResNet-101 & 91.6 & 84.0 & 88.6 & 96.0 & 94.9 & 91.0\\
        $\textrm{DiT}_{\rm BASE}$~\cite{li2022dit} & Mask R-CNN & Transformer & 93.4 & 87.1 & 92.9 & 97.3 & 96.7 & 93.5\\
        UDoc~\cite{gu2021unidoc} & Faster R-CNN & ResNet-50 & 93.9 & 88.5 & 93.7 & 97.3 & 96.4 & 93.9 \\
        $\textrm{DiT}_{\rm BASE}$~\cite{li2022dit} & Cascade R-CNN & Transformer & 94.4 & 88.9 & 94.8 & 97.6 & 96.9 & 94.5\\
        \midrule
        \bf $\textrm{LayoutLMv3}_{\rm BASE}$ (Ours) & Cascade R-CNN & Transformer & \bf 94.5 & \bf 90.6 & \bf 95.5 & \bf 97.9 & \bf 97.0 & \bf 95.1 \\
        \bottomrule
    \end{tabular}
\end{table*}

We fine-tune LayoutLMv3 on multimodal tasks on publicly available benchmarks.
Results are shown in Table~\ref{tab:sota}.

\noindent \textbf{Task I: Form and Receipt Understanding.}
Form and receipt understanding tasks require extracting and structuring forms and receipts' textual content.
The tasks are a sequence labeling problem aiming to tag each word with a label.
We predict the label of the last hidden state of each text token with a linear layer and an MLP classifier for form and receipt understanding tasks, respectively. 

We conduct experiments on the FUNSD dataset and the CORD dataset.
The \textbf{FUNSD}~\cite{jaume2019funsd} is a noisy scanned form understanding dataset sampled from the RVL-CDIP dataset~\cite{harley2015icdar}.
The FUNSD dataset contains 199 documents with comprehensive annotations for 9,707 semantic entities. We focus on the semantic entity labeling task on the FUNSD dataset to assign each semantic entity a label among ``question'', ``answer'', ``header'' or ``other''.
The training and test splits contain 149 and 50 samples, respectively.
\textbf{CORD}~\cite{park2019cord} is a receipt key information extraction dataset with 30 semantic labels defined under 4 categories.
It contains 1,000 receipts of 800 training, 100 validation, and 100 test examples. 
We use officially-provided images and OCR annotations.
We fine-tune LayoutLMv3 for 1,000 steps with a learning rate of $1e-5$ and a batch size of 16 for FUNSD, and $5e-5$ and 64 for CORD.

We report F1 scores for this task.
For the large model size, the LayoutLMv3 achieves an F1 score of 92.08 on the FUNSD dataset, which significantly outperforms the SOTA result of 85.14 provided by StructuralLM~\cite{Li2021StructuralLMSP}.
Note that LayoutLMv3 and StructuralLM use segment-level layout positions, while the other works use word-level layout positions. Using segment-level positions may benefit the semantic entity labeling task on FUNSD~\cite{Li2021StructuralLMSP}, so the two types of work are not directly comparable.
The LayoutLMv3 also achieves SOTA F1 scores on the CORD dataset for both base and large model sizes.
The results show that LayoutLMv3 can significantly benefit the text-centric form and receipt understanding tasks.

\noindent \textbf{Task II: Document Image Classification.}
The document image classification task aims to predict the category of document images.
We feed the output hidden state of the special classification token ([CLS]) into an MLP classifier to predict the class labels.

We conduct experiments on the \textbf{RVL-CDIP} dataset.
It is a subset of the IIT-CDIP collection labeled with 16 categories~\cite{harley2015icdar}. 
RVL-CDIP dataset contains 400,000 document images, among them 320,000 are training images, 40,000 are validation images, and 40,000 are test images.
We extract text and layout information using Microsoft Read API.
We fine-tune LayoutLMv3 for 20,000 steps with a batch size of 64 and a learning rate of $2e-5$.

The evaluation metric is the overall classification accuracy.
LayoutLMv3 achieves better or comparable results with a much smaller model size than previous works.
For example, compared to LayoutLMv2, LayoutLMv3 achieves an absolute improvement of 0.19\% and 0.29\% in the base model and large model size, respectively, with a much simpler image embedding (i.e., Linear vs. ResNeXt101-FPN).
The results show that our simple image embeddings can achieve desirable results on image-centric tasks.

\noindent \textbf{Task III: Document Visual Question Answering.}
Document visual question answering requires a model to take a document image and a question as input and output an answer~\cite{mathew2021docvqa}.
We formalize this task as an extractive QA problem, where the model predicts start and end positions by classifying the last hidden state of each text token with a binary classifier.

We conduct experiments on the \textbf{DocVQA} dataset, a standard dataset for visual question answering on document images~\cite{mathew2021docvqa}. 
The official partition of the DocVQA dataset consists of 10,194/1,286/1,287 images and 39,463/5,349/5,188 questions for training/validation/test set, respectively. We train our model on the training set, evaluate the model on the test set, and report results by submitting them to the official evaluation website.
We use Microsoft Read API to extract text and bounding boxes from images and use heuristics to find given answers in the extracted text as in LayoutLMv2.
We fine-tune $\textrm{LayoutLMv3}_{\rm BASE}$ for 100,000 steps with a batch size of 128, a learning rate of $3e-5$, and a warmup ratio of 0.048.
For $\textrm{LayoutLMv3}_{\rm LARGE}$, the step size, batch size, learning rate and warmup ratio are 200,000, 32, $1e-5$, and 0.1, respectively.

We report the commonly-used edit distance-based metric ANLS (also known as Average Normalized Levenshtein Similarity).
The $\textrm{LayoutLMv3}_{\rm BASE}$ improves the ANLS score of $\textrm{LayoutLMv2}_{\rm BASE}$ from 78.08 to 78.76, with much simpler image embedding (i.e., from ResNeXt101-FPN to Linear embedding).
The $\textrm{LayoutLMv3}_{\rm LARGE}$ further gains an absolute ANLS score of 4.61 over $\textrm{LayoutLMv3}_{\rm BASE}$.
The results show that LayoutLMv3 is effective for the document visual question answering task.

\begin{table*}[t]
    \centering
    \caption{\textbf{Ablation study on image embeddings and pre-training objectives} on typical text-centric tasks (form and receipt understanding on FUNSD and CORD) and image-centric tasks (document image classification on RVL-CDIP and document layout analysis on PubLayNet).
    All models were trained at $\mathrm{BASE}$ size on 1 million data for 150,000 steps with learning rate $3e-4$.
    }
    \label{tab:ablation}
    \begin{tabular}{ccclcccc}
    \toprule
    \multirow{2}{*}{\bf \#} & \bf Image & \multirow{2}{*}{\bf Parameters} & \bf Pre-training & \bf FUNSD & \bf CORD & \bf RVL-CDIP  & \bf PubLayNet \\
     & \bf Embed & & \bf Objective(s) & F1$\uparrow$ & F1$\uparrow$ & Accuracy$\uparrow$ & MAP$\uparrow$ \\
    \midrule
    1 & None & 125M & MLM & {88.64} & {96.27} & 95.33 & Not Applicable \\
    2 & Linear & 126M & MLM & 89.39 & 96.11 & 95.00 & Loss Divergence \\
    3 & Linear & 132M & MLM+MIM  & 89.19 & 96.30 & {95.42}& {94.38} \\
    4 & Linear & 133M & MLM+MIM+WPA & \textbf{89.78} &  \textbf{96.49} & \textbf{95.53} & \textbf{94.43} \\
    \bottomrule
    \end{tabular}
\end{table*}

\subsection{Fine-tuning on a Vision Task}
To demonstrate the generality of LayoutLMv3 from the multimodal domain to the visual domain, we transfer LayoutLMv3 to a \textbf{document layout analysis} task.
This task is about detecting the layouts of unstructured digital documents by providing bounding boxes and categories such as tables, figures, texts, etc.
This task helps parse the documents into a machine-readable format for downstream applications.
We model this task as an object detection problem without text embedding, which is effective in existing works~\cite{zhong2019publaynet, gu2021unidoc, li2022dit}.
We integrate the LayoutLMv3 as feature backbone in the Cascade R-CNN detector~\cite{cai2018cascade} with FPN~\cite{lin2017feature} implemented using the Detectron2~\cite{wu2019detectron2}.
We adopt the standard practice to extract single-scale features from different Transformer layers, such as layers 4, 6, 8, and 12 of the LayoutLMv3 base model. We use resolution-modifying modules to convert the single-scale features into the multiscale FPN features~\cite{ali2021xcit,li2021benchmarking,li2022dit}.

We conduct experiments on \textbf{PubLayNet} dataset~\cite{zhong2019publaynet}.
The dataset contains research paper images annotated with bounding boxes and polygonal segmentation across five document layout categories: text, title, list, figure, and table.
The official splits contain 335,703 training images, 11,245 validation images, and 11,405 test images. We train our model on the training split and evaluate our model on the validation split following standard practice~\cite{zhong2019publaynet,gu2021unidoc,li2022dit}.
We train our model for 60,000 steps using the AdamW optimizer with 1,000 warm-up steps and a weight decay of 0.05 following DiT~\cite{li2022dit}. Since LayoutLMv3 is pre-trained with inputs from both vision and language modalities, we use a larger batch size of 32 and a lower learning rate of $2e-4$ empirically.
We do not use flipping or cropping augmentation strategy in the fine-tuning stage to be consistent with our pre-training stage.
We do not use relative positions in self-attention networks as DiT.

We measure the performance using the mean average precision (MAP) @ intersection over union (IOU) [0.50:0.95] of bounding boxes and report results in Table~\ref{tab:publaynet}.
We compare with the ResNets~\cite{zhong2019publaynet, gu2021unidoc} and the concurrent vision Transformer~\cite{li2022dit} backbones.
LayoutLMv3 outperforms the other models in all metrics, achieving an overall mAP score of 95.1.
LayoutLMv3 achieves a high gain in the ``Title'' category. Since titles are typically much smaller than other categories and can be identified by their textual content, we attribute this improvement to our incorporation of language modality in pre-training LayoutLMv3.
These results demonstrate the generality and superiority of LayoutLMv3.

\begin{figure}[t]
    \centering
    \includegraphics[width=\linewidth]{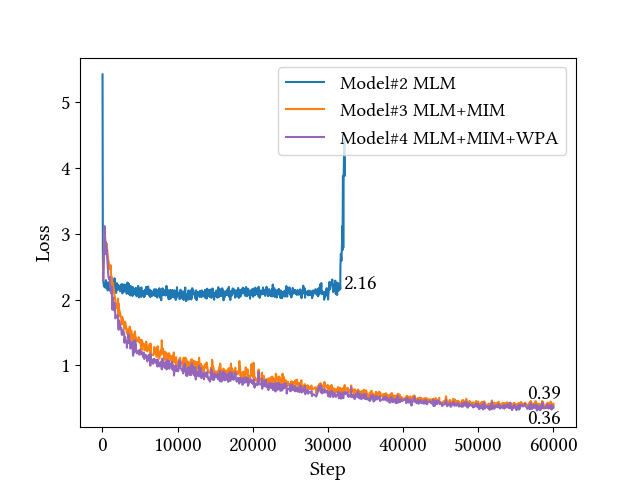}
    \caption{Loss convergence curves of fine-tuning the ablated models of  LayoutLMv3 on PubLayNet dataset.
    The loss of model \#2 did not converge.
    By incorporating the MIM objective, the loss converges normally.
    The WPA objective further decreases the loss.
    Best viewed in color.
    }
    \label{fig:loss}
\end{figure} 

\subsection{Ablation Study}\label{subsec:ablation_study}
In Table~\ref{tab:ablation} we study the effect of our image embeddings and pre-training objectives.
We first build a baseline model \#1 that uses text and layout information, pre-trained with MLM objective.
Then we use linearly projected image patches as the image embedding of the baseline model, denoted as model \#2.
We further pre-train model \#2 with MIM and WPA objectives step by step and denote the new models as \#3 and \#4, respectively.

In Figure~\ref{fig:loss}, we visualize losses of models \#2, \#3, and \#4 when fine-tuned on the PubLayNet dataset with a batch size of 16 and a learning rate of $2e-4$.
We have tried to train the model \#2 with learning rates of \{$1e-4$, $2e-4$, $4e-4$\} combined with batch sizes of \{$16$, $32$\}, but the loss of model \#2 did not converge and the mAP score on PubLayNet is near zero.

\noindent \textbf{Effect of Linear Image Embedding.}
We observe that model \#1 without image embedding has achieved good results on some tasks.
This suggests that language modality, including text and layout information, plays a vital role in document understanding.
However, the results are still unsatisfactory.
Moreover, model \#1 cannot conduct some image-centric document analysis tasks without vision modality.
For example, the vision modality is critical for the document layout analysis task on PubLayNet because bounding boxes are tightly integrated with images.
Our simple design of linear image embedding combined with appropriate pre-training objectives can consistently improve not only image-centric tasks, but also some text-centric tasks further.

\noindent \textbf{Effect of MIM pre-training objective.}
Simply concatenating linear image embedding with text embedding as input to model \#2 deteriorates performance on CORD and RVL-CDIP, while the loss on PubLayNet diverges.
We speculate that the model failed to learn meaningful visual representation on the linear patch embeddings without any pre-training objective associated with image modality.
The MIM objective mitigates this problem by preserving the image information until the last layer of the model by randomly masking out a portion of input image patches and reconstructing them in the output~\cite{kim2021vilt}.
Comparing the results of model \#3 and model \#2, the MIM objective benefits CORD and RVL-CDIP.
As simply using linear image embedding has improved FUNSD, MIM does not further contribute to FUNSD.
By incorporating the MIM objective in training, the loss converges when fine-tuning PubLayNet as shown in Figure~\ref{fig:loss}, and we obtain a desirable mAP score.
The results indicate that MIM can help regularize the training. Thus MIM is critical for vision tasks like document layout analysis on PubLayNet.

\noindent \textbf{Effect of WPA pre-training objective.}
By comparing models \#3 and \#4 in Table~\ref{tab:ablation}, we observe that the WPA objective consistently improves all tasks.
Moreover, the WPA objective decreases the loss of the vision task on PubLayNet in Figure~\ref{fig:loss}.
These results confirm the effectiveness of WPA not only in cross-modal representation learning, but also in image representation learning.

\noindent \textbf{Parameter Comparisons.}
The table shows that incorporating image embedding for a $16 \times 16$ patch projection (\#1 $\rightarrow$ \#2) introduces only 0.6M parameters.
The parameters are negligible compared to the parameters of CNN backbones (e.g., 44M for ResNet-101).
A MIM head and a WPA head introduce 6.9M and 0.6M parameters in the pre-training stage.
The parameter overhead introduced by image embedding is marginal compared to the MLM head,  which has 39.2M parameters for a text vocabulary size of 50,265.
We did not take count of the image tokenizer when calculating parameters as the tokenizer is a standalone module for generating the labels of MIM but is not integrated into the Transformer backbone.

\section{Related Work}

\textbf{Multimodal self-supervised pre-training} technique has made a rapid progress in \emph{document intelligence} due to its successful applications of document layout and image representation learning~\cite{xu2020layoutlm,xu-etal-2021-layoutlmv2,xu2021layoutxlm,pramanik2020towards,garncarek2021lambert,hong2022bros,Powalski2021GoingFB,wu2021lampret,Li2021StructuralLMSP,li2021selfdoc,Appalaraju_2021_ICCV,li2021structext,gu2021unidoc,wang2022LiLT,gu2022xylayoutlm,lee2022formnet}.
LayoutLM and following works joint layout representation learning by encoding spatial coordinates of text~\cite{xu2020layoutlm,Li2021StructuralLMSP,hong2022bros,lee2022formnet}. 
Various works then joint image representation learning by combining CNNs with Transformer~\cite{vaswani2017attention} self-attention networks.
These works either extract CNN grid features~\cite{xu-etal-2021-layoutlmv2,Appalaraju_2021_ICCV} or rely on an object detector to extract region features~\cite{xu2020layoutlm,Powalski2021GoingFB,li2021selfdoc,gu2021unidoc}, which accounts for heavy computation bottleneck or requires region supervision.
In the field of \emph{natural images vision-and-language pre-training} (VLP), research works have seen a shift from region features~\cite{tan2019lxmert,su2019vl,chen2020uniter} to grid features~\cite{huang2021seeing} to lift limitations of pre-defined object classes and region supervision.
Inspired by vision Transformer (ViT)~\cite{dosovitskiy2020vit}, there have also been recent efforts in VLP without CNNs to overcome the weakness of CNN. Still, most rely on separate self-attention networks to learn visual features; thus, their computational cost is not reduced~\cite{xue2021probing,li2021align,dou2021empirical}.
An exception is ViLT, which learns visual features with a lightweight linear layer and significantly cuts down the model size and running time~\cite{kim2021vilt}.
Inspired by ViLT, our LayoutLMv3 is the first multimodal model in Document AI that utilizes image embeddings without CNNs.

\textbf{Reconstructive pre-training objectives} revolutionized representation learning.
In \emph{NLP} research, BERT firstly proposed ``masked language modeling'' (MLM) to learn bidirectional representations and advanced the state of the arts on broad language understanding tasks~\cite{devlin2019bert}.
In the field of \emph{CV}, Masked Image Modeling (MIM) aims to learn rich visual representations via predicting masked content conditioning in visible context.
For example, ViT reconstructs the mean color of masked patches, which leads to performance gains in ImageNet classification~\cite{dosovitskiy2020vit}.
BEiT reconstructs visual tokens learned by a discrete VAE, achieving competitive results in image classification and semantic segmentation~\cite{bao2022beit}.
DiT extends BEiT to document images to document layout analysis~\cite{li2022dit}.

Inspired by MLM and MIM, researchers in the field of \emph{vision-and-language} have explored \textbf{reconstructive objectives for multimodal representation learning}.
Whereas most well-performing vision-and-language pre-training (VLP) models use the MLM proposed by BERT on text modality, they differ in their pre-training objectives for the image modality.
There are three variants of MIM corresponding to different image embeddings: masked region modeling (MRM), masked grid modeling (MGM), and masked patch modeling (MPM).
MRM has been proven to be effective in regressing original region features~\cite{tan2019lxmert,chen2020uniter,li2021selfdoc} or classifying object labels~\cite{chen2020uniter,lu2019vilbert,tan2019lxmert} for masked regions.
MGM has also been explored in the SOHO, whose objective is to predict the mapping index in a visual dictionary for masked grid features~\cite{huang2021seeing}.
For patch-level image embedding, Visual Parsing~\cite{xue2021probing} proposed to mask visual tokens according to the attention weights in their self-attention image encoder, which does not apply to simple linear image encoders.
ViLT~\cite{kim2021vilt} and METER~\cite{dou2021empirical} attempt to leverage MPM similar to ViT~\cite{dosovitskiy2020vit} and BEiT~\cite{bao2022beit}, which respectively reconstruct the mean color and discrete tokens in visual vocabularies for image patches, but resulted in degraded performance on downstream tasks.
Our LayoutLMv3 firstly demonstrates the effectiveness of MIM for linear patch image embedding.

Various \textbf{cross-modal objectives} are further developed for vision and language (VL) alignment learning in multimodal models.
Image-text matching is widely used to learn a coarse-grained VL alignment~\cite{chen2020uniter,huang2021seeing,kim2021vilt,xu-etal-2021-layoutlmv2,Appalaraju_2021_ICCV}.
To learn a fine-grained VL alignment, UNITER proposes a word-region alignment objective based on optimal transports, which calculates the minimum cost of transporting the contextualized image embeddings to word embeddings~\cite{chen2020uniter}.
ViLT extends this objective to patch-level image embeddings~\cite{kim2021vilt}.
Unlike natural images, document images imply an explicit fine-grained alignment relationship between text words and image areas.
Using this relationship, UDoc uses contrastive learning and similarity distillation to align the image and text belonging to the same area~\cite{gu2021unidoc}.
LayoutLMv2 covers some text lines in raw images and predicts whether each text token is covered~\cite{xu-etal-2021-layoutlmv2}.
In contrast, we naturally utilize the masking operations in MIM to construct aligned/unaligned pairs in an effective and unified way.

\section{Conclusion and Future Work}
In this paper, we present LayoutLMv3 to pre-train the multimodal Transformer for Document AI, which redesigns the model architecture and pre-training objectives for LayoutLM. 
Distinguishing from the existing multimodal model in Document AI, LayoutLMv3 does not rely on a pre-trained CNN or Faster R-CNN backbone to extract visual features, significantly saving parameters and eliminating region annotations.
We use unified text and image masking pre-training objectives: masked language modeling, masked image modeling, and word-patch alignment, to learn multimodal representations.
Extensive experimental results have demonstrated the generality and superiority of LayoutLMv3 for both text-centric and image-centric Document AI tasks with the simple architecture and unified objectives.
In future research, we will investigate scaling up pre-trained models
so that the models can leverage more training data 
to drive SOTA results further. In addition, we will explore few-shot and zero-shot learning capabilities to facilitate more real-world business scenarios in the Document AI industry.

\section{Acknowledgement}
We are grateful to Yiheng Xu for fruitful discussions and inspiration.
This work was supported by the NSFC (U1811461) and the Program for Guangdong Introducing Innovative and Entrepreneurial Teams under Grant NO.2016ZT06D211.


\begin{table*}[ht]
    \centering
    \caption{\textbf{Visual information extraction in Chinese} F1 score on the EPHOIE test set.}
    \label{tab:EPHOIE}
    \resizebox{\textwidth}{!}
    {
    \begin{tabular}{llllllllllll}
        \toprule
        \multicolumn{1}{l}{\bf Model} & \bf Subject & \bf Test Time & \bf Name & \bf School & \bf \#Examination & \bf \#Seat & \bf Class & \bf \#Student & \bf Grade & \bf Score & \bf Mean \\
        \midrule
        BiLSTM+CRF~\cite{lample2016neural} & 98.51 & 100.0 & 98.87 & 98.80 & 75.86 & 72.73 & 94.04 & 84.44 & 98.18 & 69.57 & 89.10 \\
        GCN-based~\cite{liu2019graph} & 98.18 & 100.0 & 99.52 & \bf 100.0 & 88.17 & 86.00 & 97.39 & 80.00 & 94.44 & 81.82 & 92.55 \\
        GraphIE~\cite{qian2019graph} & 94.00 & 100.0 & 95.84 & 97.06 & 82.19 & 84.44 & 93.07 & 85.33 & 94.44 & 76.19 & 90.26 \\
        TRIE~\cite{zhang2020trie} & 98.79 & 100.0 & 99.46 & 99.64 & 88.64 & 85.92 & 97.94 & 84.32 & 97.02 & 80.39 & 93.21 \\
        VIES~\cite{wang2021towards} & \bf 99.39 & 100.0 &  99.67 & 99.28 & 91.81 & 88.73 & \bf 99.29 & 89.47 & 98.35 & 86.27 & 95.23 \\
        StrucTexT~\cite{li2021structext} & 99.25 & 100.0 & 99.47 & 99.83 & 97.98 & 95.43 & 98.29 & 97.33 & \bf 99.25 & 93.73 & 97.95 \\
        \midrule
        \bf $\textrm{LayoutLMv3-Chinese}_{\rm BASE}$ (Ours) & 98.99 & \bf 100.0 & \bf 99.77 & 99.20 & \bf 100.0 & \bf 100.0 & 98.82 & \bf 99.78 & 98.31 & \bf 97.27 & \bf 99.21 \\
        \bottomrule
    \end{tabular}
    }
\end{table*}

\newpage\appendix
\section{Appendix}

\subsection{LayoutLMv3 in Chinese}

\noindent \textbf{Pre-training LayoutLMv3 in Chinese.}
To demonstrate the effectiveness of LayoutLMv3 in not only English but also in the Chinese language, we pre-train a LayoutLMv3-Chinese model in base size. It is trained on 50 million document pages in Chinese.
We collect large-scale Chinese documents by downloading publicly available digital-born documents and following the principles of Common Crawl (https://commoncrawl.org/) to process these documents.
For the multimodal Transformer encoder along with the text embedding layer, LayoutLMv3-Chinese is initialized from the pre-trained weights of XLM-R~\cite{conneau2020unsupervised}.
We randomly initialized the rest model parameters.
Other training setting is the same as LayoutLMv3.

\noindent \textbf{Fine-tuning on Visual Information Extraction.}
The visual information extraction (VIE) requires extracting key information from document images.
The task is a sequence labeling problem aiming to tag each word with a pre-defined label.
We predict the label of the last hidden state of each text token with a linear layer. 

We conduct experiments on the EPHOIE dataset.
The \textbf{EPHOIE}~\cite{wang2021towards} is a visual information extraction dataset consisting of examination paper heads with diverse layouts and backgrounds.
It contains 1,494 images with comprehensive annotations for 15,771 Chinese text instances.
We focus on a token-level entity labeling task on the EPHOIE dataset to assign each character a label among ten pre-defined categories.
The training and test sets contain 1,183 and 311 images, respectively.
We fine-tune LayoutLMv3-Chinese for 100 epochs. The batch size is 16, and the learning rate is $5e-5$ with linear warmup over the first epoch.

We report F1 scores for this task and report results in Table~\ref{tab:EPHOIE}.
The LayoutLMv3-Chinese shows superior performance on most metrics and achieves a SOTA mean F1 score of 99.21\%. The results show that LayoutLMv3 significantly benefits the VIE task in Chinese.

\bibliographystyle{ACM-Reference-Format}
\balance
\bibliography{ref}

\end{document}